%% file: main.tex
\documentclass{article}

\usepackage{authblk}

\usepackage{amsmath,amssymb,amsfonts}
\usepackage{algorithmic}
\usepackage{graphicx}
\usepackage[T1]{fontenc}
\usepackage{multirow}
\usepackage{multicol}
\usepackage{array}
\usepackage{url}
\usepackage{booktabs}

\usepackage{verbatim}

\title{Compressing Sentence Representation with Maximum Coding Rate Reduction}

\author[1]{Domagoj Ševerdija\thanks{\{dseverdi, tprusina, ajovanov, lborozan, jmaltar, domagoj\}@mathos.hr}}
\author[1]{Tomislav Prusina\protect\footnotemark[1]}
\author[1]{Antonio Jovanović\protect\footnotemark[1]}
\author[1]{Luka Borozan\protect\footnotemark[1]}
\author[1]{Jurica Maltar\protect\footnotemark[1]}
\author[1]{Domagoj Matijević\protect\footnotemark[1]}

\affil[1]{Department of Mathematics, University J.~J.~Strossmayer of Osijek, Croatia}

\begin{document}

\maketitle

\begin{abstract}
In most natural language inference problems, sentence representation is needed for semantic retrieval tasks. In recent years, pre-trained large language models have been quite effective for computing such representations. These models produce high-dimensional sentence embeddings. An evident performance gap between large and small models exists in practice. Hence, due to space and time hardware limitations, there is a need to attain comparable results when 
using the smaller model, which is usually a distilled version of the large language model. In this paper, we assess the model distillation of the sentence representation model Sentence-BERT by augmenting the pre-trained distilled model with a projection layer additionally learned 
on the Maximum Coding Rate Reduction ($\textrm{MCR}^2$) objective, a novel approach developed for general purpose manifold clustering.

We demonstrate that the new language model with reduced complexity and sentence embedding size can achieve comparable results on semantic retrieval benchmarks.

\end{abstract}


\section{Introduction}
Dense vector representations of words, or word embeddings, form the backbone of most NLP applications and can be constructed using context-free (see \cite{10.5555/944919.944966}, \cite{Mikolov13}, \cite{pennington-etal-2014-glove}) or contextualized  methods (see \cite{devlin-etal-2019-bert}, \cite{peters-etal-2018-deep} for more details). In practice, few NLP applications often benefit from having sentence or document representations in addition to word embeddings. In most cases, one can use the weighted average (aka pooling) over some or all of the word embeddings from a sentence or document. Although it disregards word order while pooling, this approach has shown to be reasonably performant \cite{https://doi.org/10.48550/arxiv.1806.04713}. Pre-trained language models like BERT have shown success on many NLP tasks through fine-tuning. Unfortunately, using contextualized word vectors from these models as a sentence representation is significantly inferior in terms of semantic textual similarity  compared to approaches when one uses non-contextualized word vectors, which are trained with a much simpler model (see  \cite{reimers_sentence-bert_2019} for more details). Therefore, more sophisticated methods were derived to find efficient and performant universal sentence encoders. Reimers et al. in \cite{reimers_sentence-bert_2019} developed the Sentence-BERT model by fine-tuning pre-trained BERT architecture on sentence pair scoring tasks using a Siamese architecture to learn better sentence representations, showing much improvement in downstream NLP tasks.
Their approach ended up with a relatively large model size (hundreds of millions to billions of parameters) and sentence embedding dimension 768, a relatively large number for efficient search and retrieval operations over databases. 
In this paper, we focus on reducing the dimensionality of sentence embeddings up to 50\%-70\%  while still achieving comparable results across the board of NLP benchmarks. This opens up a possibility of deploying AI models on smaller-scale computer systems like embedded systems.

\subsection{Related Work}
Following the distributional hypothesis, Mikolov et al. in \cite{Mikolov13} showed that computing dense vectors of lower dimension for word embeddings give interesting mathematical properties of words. Inspired by the same idea, Kiros et al. \cite{10.5555/2969442.2969607} and Lee et al.\cite{DBLP:journals/corr/abs-1803-02893} tried to derive a model which predicts surrounding sentences. Sent2Vec \cite{Pagliardini_2018} generates context-free sentence embeddings as averages of word vectors and $n$-gram vectors (similar to FastText \cite{bojanowski2016enriching} for words).  Conneau et al.\cite{conneau-etal-2017-supervised} computed contextualized sentence embeddings using a BiLSTM Siamese network that was fine-tuned on pairs of semantically similar sentences. This approach was extended to fine-tuning pre-trained language models like BERT in \cite{reimers_sentence-bert_2019}. Recently, Gao et al. \cite{gao-etal-2021-simcse} improved this approach by suggesting a contrastive learning method and achieved state-of-the-art results. Projecting sentence embeddings to lower dimensions was motivated by projecting word vectors. In most cases, PCA methods gave surprisingly good results and even retrofitted the word vectors in such a way that it made vectors more isotropic which had a good impact on NLP benchmarks. Li et al. \cite{li-etal-2020-sentence} showed that this anomaly is also apparent in sentence vectors and gave a normalizing flow method to retrofit such vectors. Recent work of \cite{yu_learning_2020} introduced Maximum Coding Rate Reduction (MCR$^2$), a novel learning objective that enables for learning a subspace representation given the clustering\footnote{The formal definition of MCR$^2$ will be given in Section~\ref{sec:method}.}.  They also demonstrated how to extend the approach to the problem of unsupervised clustering. 

\subsection{Our contribution} 
We use a pre-trained sentence embedding model like Sentence-BERT (SBERT) as a sentence encoder and train a non-linear mapper atop the encoder using a Maximal Coding Rate Reduction as a training objective for learning discriminative low-dimensional structures that preserve all the essential information encoded into the high-dimensional data. This approach allows for more robust training than standard training objectives like cross-entropy and produces clusters in the embedding space. The main contribution of our paper is a sentence embedding compression technique that  achieves comparable results with smaller sentence embedding sizes on semantic NLP benchmarks compared to the baseline sentence encoder.

The paper is organized as follows. In Section~\ref{sec:method} we describe Maximum Rate Coding Reduction training objective for computing subspace embedding space. Furthermore, SBERT architecture is described as a sentence encoder followed by a definition of the projection layer. In Section~\ref{sec:experiments} we experimentally evaluate our method and conclude with a results discussion.

\section{Method}\label{sec:method}
For a given set of sentences $S$ and for each sentence
$$\left(word_1, word_2, \dots, word_{n_i}\right) \in S$$ our task is to construct a lower dimensional embedding $z_i \in \mathbb{R}^d$ that contains important semantic information characteristic for that sentence.
Our idea is to extend SBERT and from it's embedding compute a small projector to reduce the dimension, i.e. given the set of SBERT's embeddings $Z \in \mathbb{R}^{d\times n}$ of the dataset $S$, find a $\hat{Z} \in \mathbb{R}^{\hat{d}\times n}$ that retains semantic information extracted by SBERT.

\subsection{Learning a subspace representation with MCR$^2$}
Using the idea from Li et al.~\cite{li_neural_2022} we aim to minimize the angle between similar sentences
and maximize the entropy of the whole dataset.
For two representations $\hat{z}_1, \hat{z}_2 \in \mathbb{R}^d$ of two sentences we measure how similar they are by cosine similarity
\begin{equation*}
    D\left(\hat{z}_1, \hat{z}_2\right) = \frac{\cos \left(\hat{z}^{\top}_1\hat{z}_2\right)}{\|\hat{z}_1\|_2\|\hat{z}_2\|_2}.
\end{equation*}
For two sets $\hat{Z}_1, \hat{Z}_2 \in \mathbb{R}^{d \times b}$ we define this function as
\begin{equation}\label{eq:sim}
    D(\hat{Z}_1, \hat{Z}_2) = \frac{1}{b} \sum \limits_{i = 1}^b D\left(\hat{z}_{1,i}, \hat{z}_{2,i}\right)
\end{equation}
where $\hat{z}_{1,i}$ is the $i$-th element of $\hat{Z}_1$ and
$\hat{z}_{2,i}$ is the $i$-th element of $\hat{Z}_2$.
Given pairs of similar sentences we want them to have the $D$ score as large 
as possible.

For a set of representations $\hat{Z} \in \mathbb{R}^{d \times n}$
with $n$ elements, its entropy is defined as
\begin{equation}\label{eq:big_R}
    R_{\varepsilon}(\hat{Z}) = \frac{1}{2}\log \det \left(I + \frac{d}{n\varepsilon^2} \hat{Z}\hat{Z}^{\top}\right)
\end{equation}
for a given parameter $\varepsilon$ and identity matrix $I$. This function is approximately the Shannon coding rate function for multivariate Gaussian distribution given average distortion $\varepsilon$~\cite{10.5555/1146355}.
Maximizing \eqref{eq:big_R} we maximize the volume of the ball in which the embeddings are packed.
The theory behind this is well over the scope of this paper. It is given in the paper by Ma et al.~\cite{MCR2-volume-Proof} where they explore rate distortion, $\varepsilon$-ball packing and lossy encoding with normally distributed data.
By optimizing it in parallel with \eqref{eq:sim} we try to distance each sentence from others, except for the similar pairs that we try to keep close. Additionally, given cluster assignments, we can measure the entropy of each cluster with
\begin{equation}\label{eq:sum_R}
    R_\varepsilon\left(\hat{Z}, \Pi_k\right) = \frac{n_k}{2n}\log \det \left(I + \frac{d}{n_k\varepsilon^2} \hat{Z}\Pi_k\hat{Z}^{\top}\right)
\end{equation}
where $\Pi_k$ is a diagonal matrix with $i$-th entry being 1 if the $i$-th sentence belongs to cluster $k$, otherwise 0, and $n_k = \text{tr}\left(\Pi_k\right)$, trace of matrix $\Pi_k$, i.e. number of points in this cluster.
Combining functions \eqref{eq:sim}, \eqref{eq:big_R} and \eqref{eq:sum_R} into one we get the MRC$^2$ loss function defined with
\begin{equation}\label{eq:MCR2}
    L(\hat{Z}, \Pi) = - R_\varepsilon(\hat{Z})
    +\sum \limits_{i = 1}^k R_\varepsilon(\hat{Z}, \Pi_i)
    -\lambda D(\hat{Z}_1, \hat{Z}_2)
\end{equation}
for some hyperparameter $\lambda$ and pairs of similar sentences respectively divided into two sets $\hat{Z}_1, \hat{Z}_2$.
$\Pi$ denoted in $L(\hat{Z}, \Pi)$ is the clustering of data given by the user or learned by the architecture.
The choice of $\lambda$ depends on how close we want to keep similar sentences in our projection.
For larger values of $\lambda$ the network focuses on collapsing similar pairs into the same vector which,
if one is not careful enough, can lead to collapsing all vectors into one.
For smaller values of $\lambda$ the network has more freedom to decide which vector embeddings to keep close. This, 
on the other hand, can lead to an unwanted vector representation that tends to maximally distance vectors from each other.
By minimizing \eqref{eq:MCR2} we
\begin{itemize}
 \item maximize the volume of all embeddings, $R_\varepsilon(\hat{Z})$,
 \item minimize the volume of each cluster, $\sum \limits_{i=1}^k R_\varepsilon(\hat{Z}, \Pi_i)$,
 \item maximize the cosine similarity of pairs of similar sentences, $\lambda D(\hat{Z}_1, \hat{Z}_2)$.
\end{itemize}
The consequence of this is that after the minimization we have an embedding in which different clusters are orthogonal to each other
(see \cite{yu_learning_2020} for more details), i.e.
\begin{equation}\label{eq:ortho}
    i\neq j \implies \hat{Z}_i\hat{Z}^{\top}_j = 0.
\end{equation}

\subsection{Architecture}

\begin{figure*}[h]
    \centering
    \includegraphics[scale=0.45]{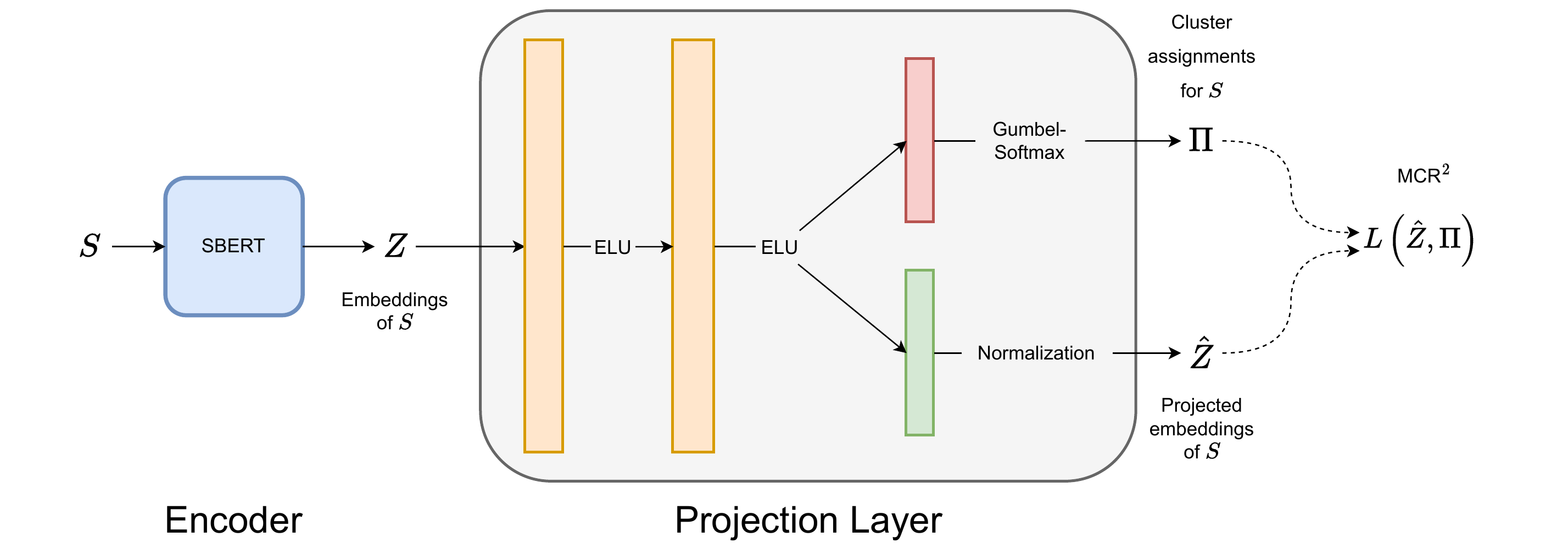}
    \caption{The overall architecture}
    \label{fig:model}
\end{figure*}

Our model receives as input a batch of sentences $S$, encodes a sentence representations $Z$ and outputs projected sentence representations $\hat{Z}$ together with cluster assignments $\Pi$ for $S$. The overall architecture is shown in Fig.~\ref{fig:model}.

\subsubsection{Sentence encoder}
BERT \cite{devlin-etal-2019-bert} and its variants has set a new state-of-the art performance on sentence-pair regression and classification tasks. Unfortunately, it requires that both sentences are fed into network causing a computation overhead which renders simple tasks like finding similar sentence pairs in large datasets a costly procedure. Therefore, SBERT \cite{reimers_sentence-bert_2019} is a modification of the BERT network which uses siamese network that is able to derive semantically meaningful sentence representations. The model consists of BERT as a pre-trained encoder, a pooling layer that computes sentence representation as an average of hidden states from the  last layer of BERT.  SBERT is trained on the combination of the SNLI \cite{bowman-etal-2015-large} and MultiNLI \cite{williams-etal-2018-broad} datasets.

\subsubsection{Projection layer}
Following Li et al.~\cite{li_neural_2022} we use above mentioned SBERT as a backbone and two last linear heads used to produce features and cluster logits. Features given by the first head are additionally normalized to unit sphere and the clusters are learned from the given pairs of similar sentences.
The whole architecture is described in Fig.~\ref{fig:model} where blue denotes the SBERT model and gray denotes a feed forward neural network that we call a projection layer. In this projection layer we have two heads. The first head colored in red is a single linear layer that collects information about the clusters and applies Gumbel-Softmax~\cite{Gumbel-Softmax}. The second head colored in green is again a single linear layer that outputs features which are in turn normalized to zero mean and unit variance. The ELU activation function is used due to its good properties \cite{https://doi.org/10.48550/arxiv.1511.07289}. 

\section{Experiments} \label{sec:experiments}

We trained our model on StackExchange duplicate questions as title/title pairs, used from CQADupStack \cite{hoogeveen2015}.
The pipeline from Sentence Transformers\footnote{\url{https://github.com/UKPLab/sentence-transformers}} for SBERT and projection layer was used with default settings, 256 batch size and 50 epochs. The backbone {\it all-mpnet-base-v2} and distilled model {\it all-MiniLM-L6-v2}  \cite{reimers-2020-multilingual-sentence-bert}  as pre-trained SBERT were frozen and the only trained part was the projection layer. We refer to the former as MPNET and the latter as MiniLM. Hyperparameter $\lambda$ from equation \eqref{eq:MCR2} was set to $2000$ for dimensions $50$ and $100$, and to $4000$ for all other dimensions.  Our model is evaluated on several downstream NLP tasks. First of all, we test our model on those benchmarks that can include clustering, namely, semantic retrieval tasks.  We also show that computed low-dimension sentence representations behave reasonably well on other semantic benchmarks. The sizes of these dimensions are motivated by experimental observation of suitable word vector sizes from \cite{patel-bhattacharyya-2017-towards} and \cite{li-etal-2020-sentence} in which a connection between word vectors and sentence embeddings is established. For downstream NLP tasks such as standard textual similarity, sentiment analysis and question-type classification tasks we use available datasets from SentEval evaluation toolkit \cite{conneau-kiela-2018-senteval} for sentence embeddings. See \cite{conneau-kiela-2018-senteval} and references therein for dataset descriptions. 

All our experiments were evaluated on AMD Ryzen Threadripper 3990X 64-Core Processor @ 4.3GHz, Nvidia GeForce RTX 3090 GPU, CUDA 11.6 with PyTorch implementation 1.9.1.

\input{gfx//results.tex}


\begin{figure*}[h!]
    \centering
    \includegraphics[scale=.6]{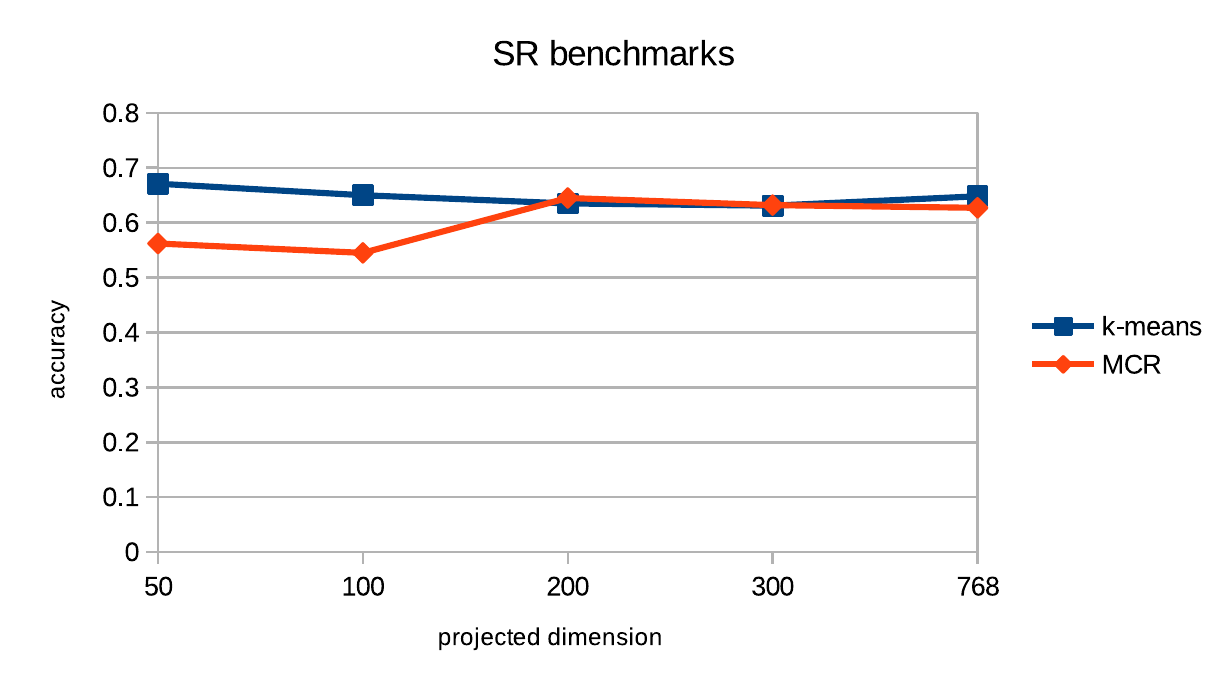} \quad
    \includegraphics[scale=.6]{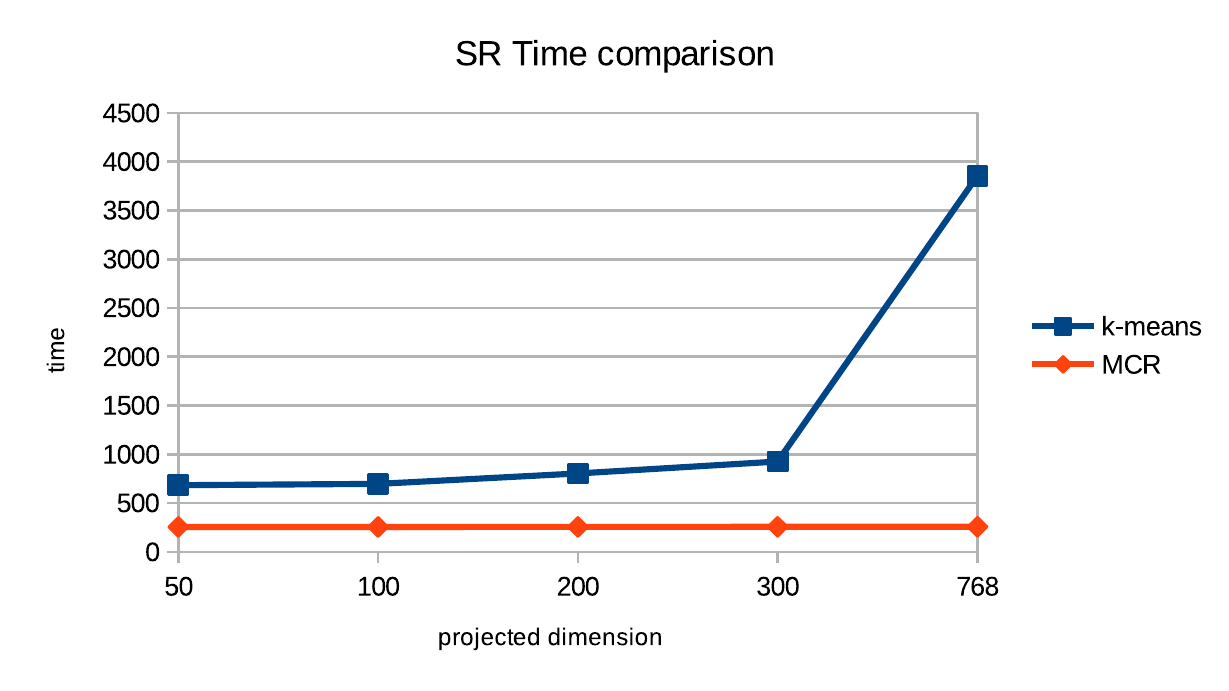} 
    \caption{\small Performance comparison on SR task}
    \label{fig:performance comparison}
\end{figure*}

\begin{figure*}[h!]
    \centering
    \includegraphics[scale=.5]{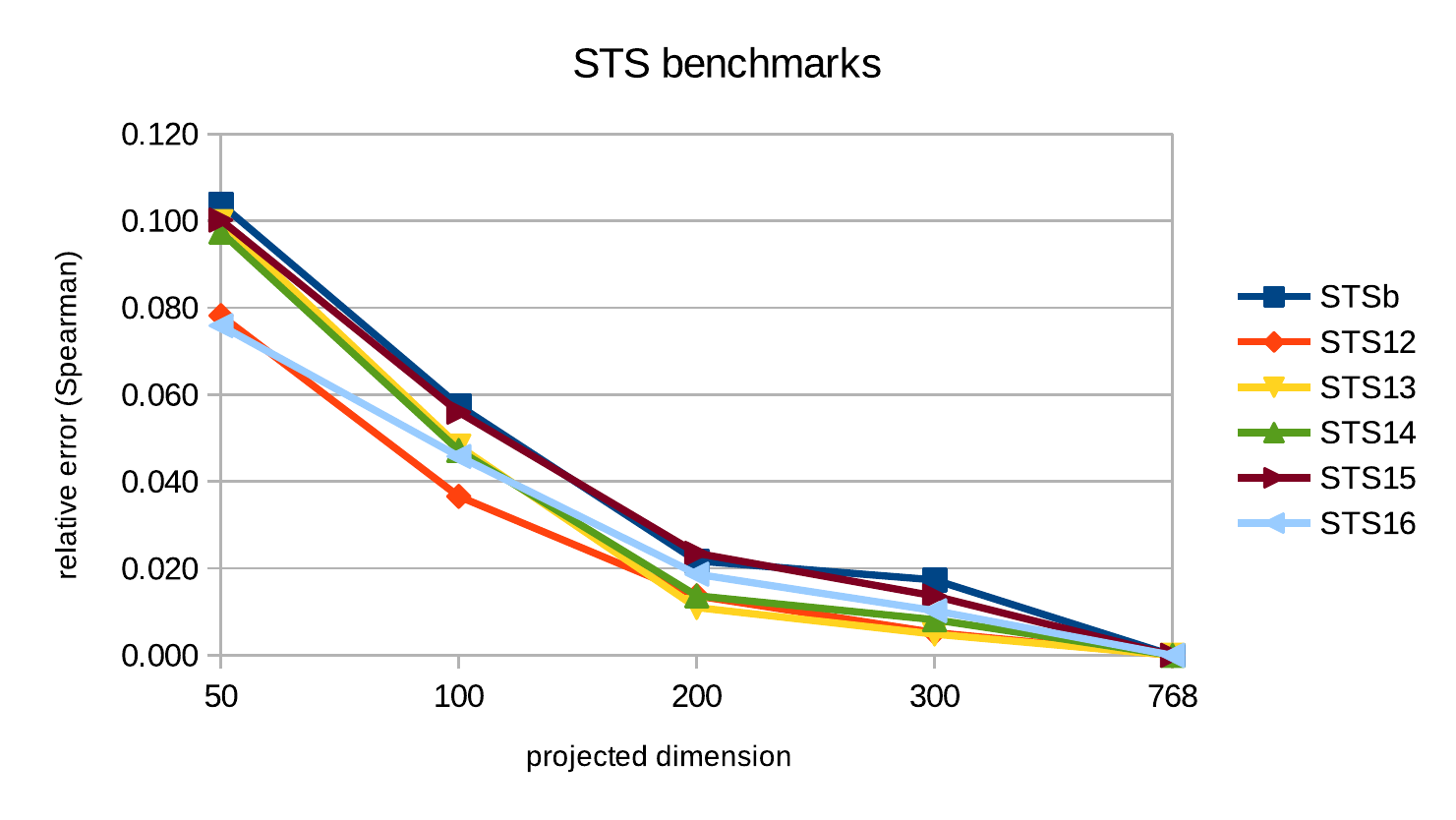} \quad
    \includegraphics[scale=.5]{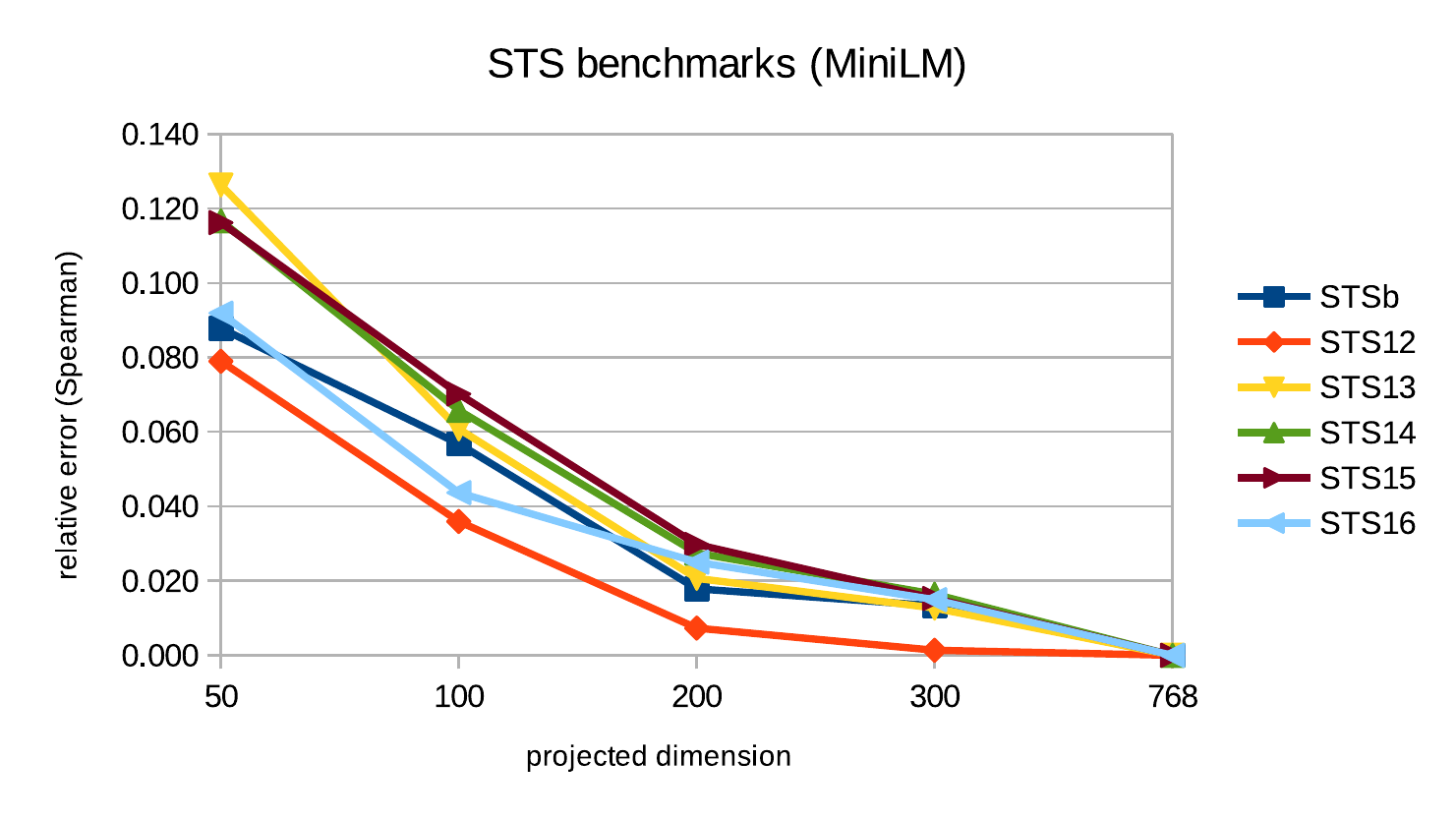} \\
    \includegraphics[scale=.5]{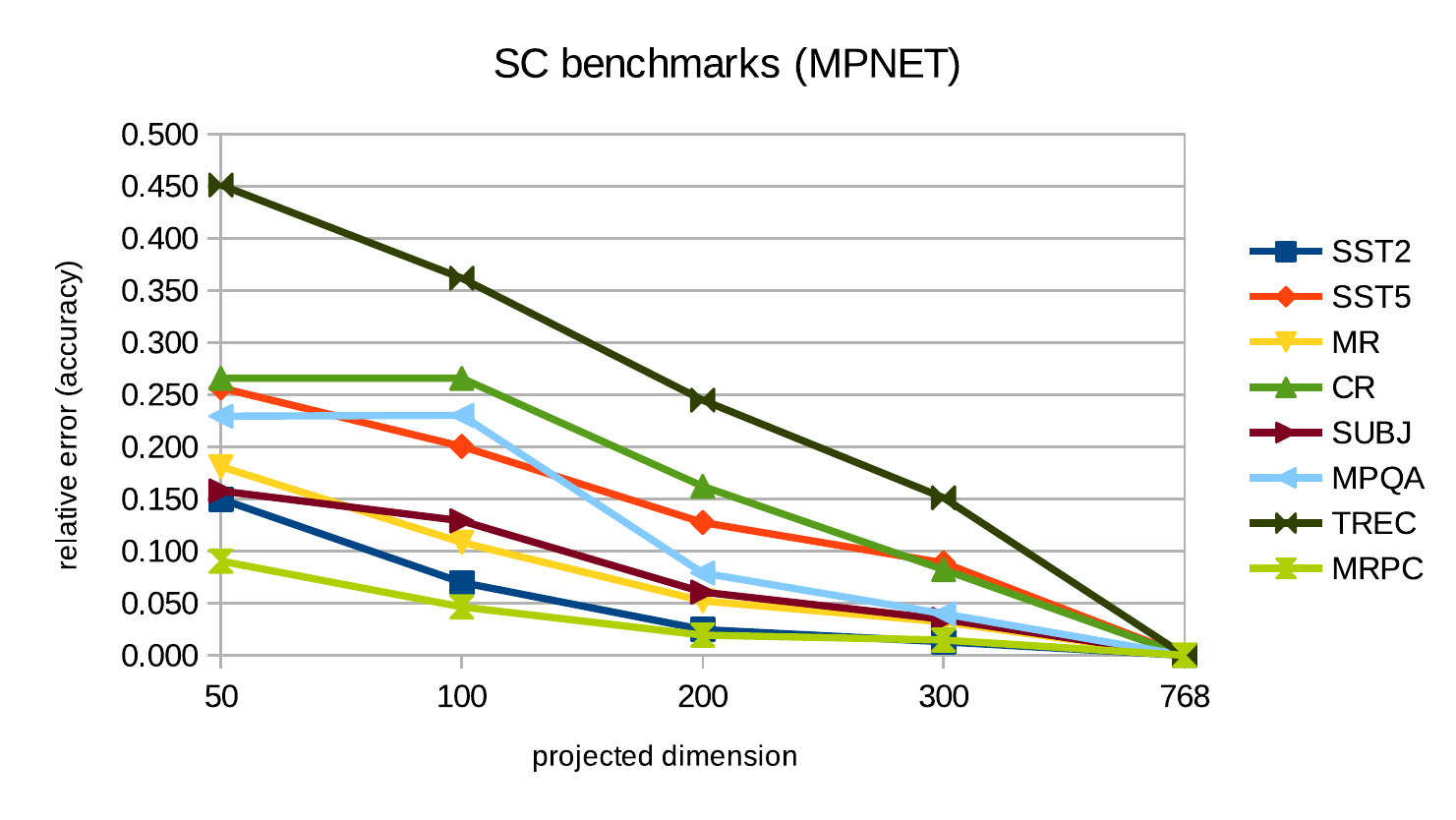} \quad
    \includegraphics[scale=.5]{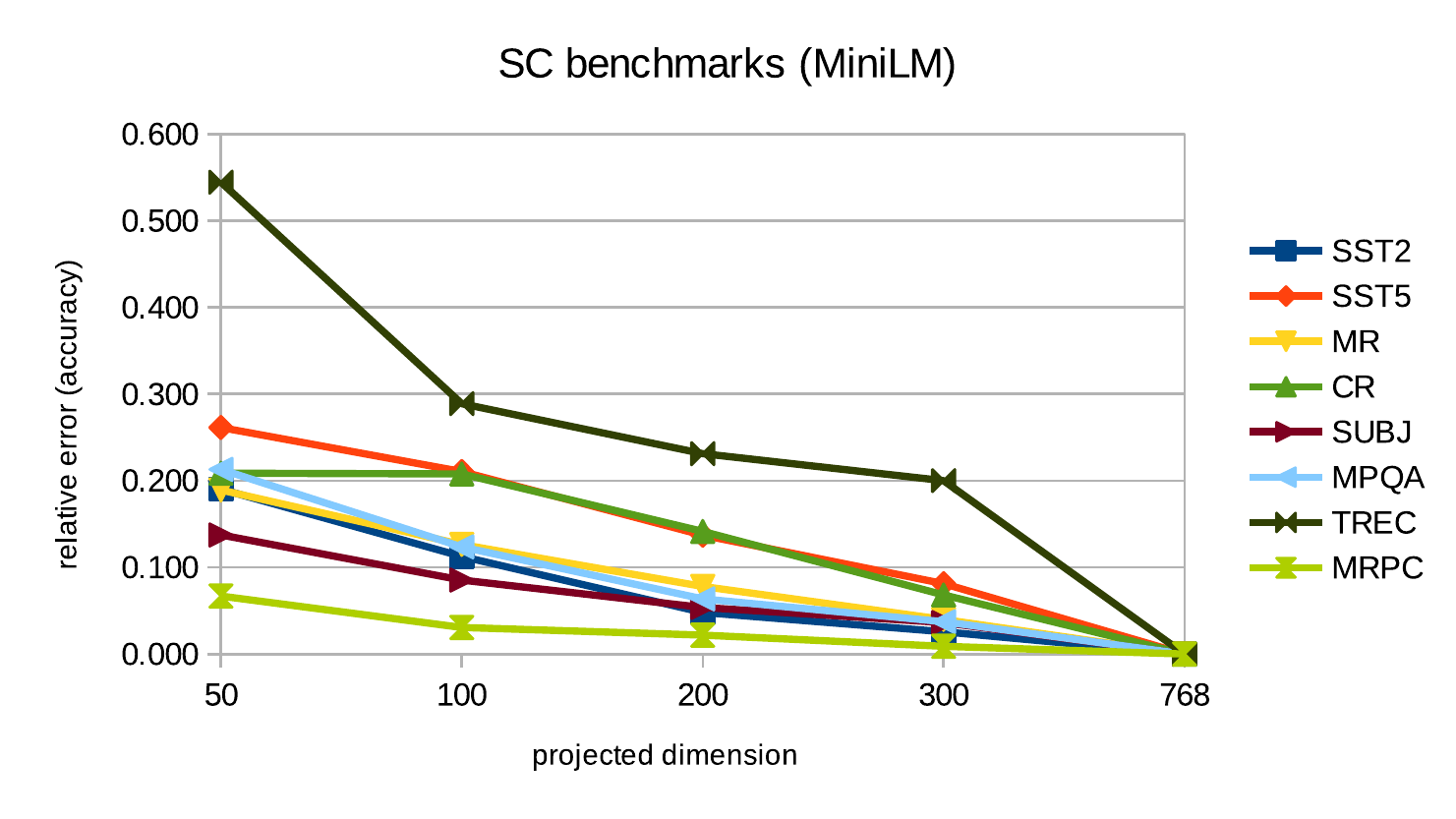}
    \caption{\small Relative error in STS and SC benchmarks}
    \label{fig:relative error}
\end{figure*}


\subsection{Semantic Retrieval (SR) Task}
The semantic retrieval (SR) task is to find all sentences in the retrieval corpus that are semantically similar to the query sentence. The basic framework is to compute sentence embeddings for the retrieval corpus and the query sentence. The goal is to find closest points in retrieval corpus embedding space to the query. Sometimes, to speed up the process \cite{johnson2019billion}, one can cluster sentences in the retrieval corpus embedding space into $k$ clusters and use query sentence to find the closest cluster of sentences. The Quora Duplicate Question Dataset\footnote{\url{https://www.kaggle.com/datasets/sambit7/first-quora-dataset}} is used to evaluate our method. This dataset consists of 500k sentences with over 400k annotated question pairs if they are duplicates or not.

\subsection{Semantic Textual Similarity (STS) Task}
One of the baseline benchmarks in natural language processing is the semantic textual similarity (STS) task that qualitatively assesses the semantic similarity between two sentences (i.e., text snippets). Our model is evaluated by computing cosine similarity between sentence pair embeddings on standard STS tasks: STS 2012-2016 and STS Benchmark available in SentEval. These datasets were labeled between 0 and 5 scores indicating the semantic relatedness of sentence pairs. Evaluation on these datasets is conducted using Spearman rank correlation which measures the correlation quality between calculated and human labeled similarity. It is valued from -1 and 1 which will be high if the ranks of predicted similarities and human labels are similar. 

\subsection{Sentence classification (SC) Task}

Sentiment classification tasks involve assigning a score for a sentiment of a snippet of text. It is formulated as a classification of text into two or more sentiment classes, namely negative, positive or neutral, or something in-between. Datasets SST, SUBJ, CR, MR are typical benchmarks for sentiment analysis.  Moreover, another example of a sentence classification task is to assign a question type for a question, like in the TREC task. In the paraphrase detection problem (like MRPC), one must classify if one sentence is a paraphrase of the other. MPQA dataset is an example of opinion classification task. The performance metric for these benchmarks is given as accuracy. All these datasets are available in SentEval toolkit.

\section{Results}

This section compares our method as a clustering and compression algorithm, respectively. In Table~\ref{tab:SR_results}, we compare how our clustering competes with $k$-means\footnote{implemented in \emph{scikit-learn} Python package} algorithm and report time performance (needed time for encoding vectors, clustering, and total time) In the second part, we test our compression against semantic relatedness tasks. The results are reported in Tables~\ref{tab:SR_results}, \ref{tab:STS_results} and \ref{tab:SC_results}. Model names in these tables are structured as follows: the sBERT pretrained model name, the abrreviation MCR indicates that MCR$^2$ is used as a projection, the number followed is a projection dimension, and optionally if $k$-means is used. The number given in parenthesis is the default embedding size.

\paragraph{Results on SR tasks} We evaluate our projection layer capacity of clustering against $k$-means clustering in retrieval space of sentence embeddings. The query sentence is assigned to a cluster of semantically related sentences, and we compare whether the ground truth duplicate belongs to that cluster, reported as an accuracy score. In all our experiments, the number of clusters was up to 128 (chosen empirically). In Table~\ref{tab:SR_results} and Fig.~\ref{fig:performance comparison}, accuracy scores and overall time for computation (i.e., encoding of sentence embeddings and clustering) depending on embedding size and type of model (MCR$^2$ with implicit clustering or $k$-means) are presented. Our method is comparable to $k$-means algorithm down to a certain dimension. On dimensions less than 200 $k$-means performed slightly better due to the fact that we did not put much effort  into finding suitable $\lambda$ values (suggested values of $\lambda$ are from \cite{li_neural_2022}). Our method computes clusters during inference which is much faster comparing to using $k$-means. Also worth noting, our projection layer used as a non-linear mapper in the original space (i.e., without any dimensionality reduction) retrofitted the sentence embeddings, enabling faster convergence of $k$-means algorithm (shown in the last row of Table~\ref{tab:SR_results}).

\paragraph{Results on STS tasks} Table~\ref{tab:STS_results} presents results for baseline (MPNET) and distilled model (MiniLM) coupled with the projection layer (MCR) with  various embedding sizes. As one can see, a relative error of up to 13\% in Spearman rank correlation is incurred if the sentence embedding dimension is as low as 6\% of the original sentence embedding size. We conclude that due to the projection layer's ability to preserve cosine distance in lower-dimensional space, the neighborhood of points is preserved, resulting in less performance degradation. This trend is visible on all STS benchmarks with both models. The relative error in Spearman rank correlation coefficient with respect to projection dimension is shown in the first row of Fig.~\ref{fig:relative error} for both the baseline and distilled model.

\paragraph{Results on SC tasks} As seen in Table~\ref{tab:SC_results}, it is observable that per-sentence classification problems like SST2 and MRPC have less performance degradation than per-token sentence classification problems like MPQA and per-sentence multi-classification problems like TREC, respectively. This is because fine-grained semantics for such tasks could not be preserved as much during projection. In the worst case, the performance degradation went up to 45\% for the baseline model and up to 60\% for the distilled model, respectively, at 6\% of the original embedding size. The relative error in accuracy with respect to projected dimension is shown in the second row of Fig.~\ref{fig:relative error} for both the baseline and distilled model.

\section{Conclusion}
In  this  paper,  we  demonstrated  how  MCR$^2$ technique could  be  used  to  obtain  lower-dimension  embeddings  of sentence  representation  for  fast  semantic  retrieval  tasks up to 70$\%$ of its original size. Also,  we  argued  that  these  embeddings  are  comparable with  SBERT  results  on  standard  semantic  NLP  benchmarks. Due to the projection layer’s ability to cluster data, we were able to cluster our sentences without any extra time cost and further reduce the representation of sentences to a reasonable dimension size without significant loss of the important semantic features. We hope our approach gives new insights for possible applications in deploying AI models in smaller-scale computer systems. 

\bibliographystyle{plain}
\bibliography{main}

\end{document}

%% file: gfx/results.tex
\begin{table*}[!h]
\scriptsize
    \caption{\small For Semantic Retrieval (SR) tasks the \texttt{all-mpnet-base-v2} SBERT model with MCR$^2$ projection to dimension 200 achieves best accuracy without no additional time for clustering like in the same setup with $k$-means (denoted with $^{*}$). Clustering backbone sentence embeddings from SBERT with $k$-means (denoted with $^{**}$) took almost an hour.}
    \label{tab:SR_results}
    \centering
\begin{tabular}{lr|lll}
\toprule
                                        &           &          \multicolumn{3}{c}{\sc Time} \\
                                   model &  accuracy & encoding & clustering &    total \\
\midrule
 all-mpnet-base-v2 + MCR50  &     0.562 & 00:04:15 &      -      & 00:04:15 \\
all-mpnet-base-v2 + MCR100  &     0.545 & 00:04:15 &      -     & 00:04:15 \\
\textbf{all-mpnet-base-v2 + MCR200}  &     0.645 & 00:04:16 &      -     & 00:04:16 \\
all-mpnet-base-v2 + MCR300  &     0.632 & 00:04:17 &      -     & 00:04:17 \\
    \midrule
    
all-mpnet-base-v2 + MCR50 + kmeans &     0.671 & 00:04:15 &   00:07:08 & 00:11:23 \\
all-mpnet-base-v2 + MCR100 + kmeans &     0.650 & 00:04:15 &   00:07:22 & 00:11:37 \\
all-mpnet-base-v2 + MCR200 + kmeans$^{*}$ &     0.635 & 00:04:16 &   00:09:08 & 00:13:24 \\
all-mpnet-base-v2 + MCR300 + kmeans &     0.631 & 00:04:17 &   00:11:09 & 00:15:26 \\
all-mpnet-base-v2 + kmeans (768)$^{**}$ &     0.648 & 00:04:17 &   00:59:57 & 01:04:12 \\
all-mpnet-base-v2 + MCR768 + kmeans 	&  	0.630 	&	00:04:17 &	00:18:15 & 	00:22:32	\\
\bottomrule
\end{tabular}
\end{table*}

\begin{table*}[!h]
\scriptsize
    \caption{\small For Semantic Textual Similarity (STS) tasks backbone model achieves the best results (bolded) for Spearman rank correlation coefficient on multiple benchmarks, although we observe comparable results of our method compared to the backbone and distilled model \texttt{all-MiniLM-L6-v2}.}
    \label{tab:STS_results}
    \centering
\begin{tabular}{lrrrrrr}
\toprule
                      model &  STSb &  STS12 &  STS13 &  STS14 &  STS15 &  STS16 \\
\midrule
 all-mpnet-base-v2 + MCR50 & 0.749 &  0.666 &  0.739 &  0.713 &  0.754 &  0.768 \\
all-mpnet-base-v2 + MCR100 & 0.788 &  0.696 &  0.782 &  0.753 &  0.791 &  0.793 \\
all-mpnet-base-v2 + MCR200 & 0.818 &  0.712 &  0.812 &  0.779 &  0.819 &  0.816 \\
all-mpnet-base-v2 + MCR300 & 0.821 &  0.718 &  0.817 &  0.783 &  0.827 &  0.823 \\
\textbf{all-mpnet-base-v2 (768)} & 0.836 &  0.722 &  0.821 &  0.790 &  0.838 &  0.831 \\
\midrule
  all-MiniLM-L6-v2 + MCR50 & 0.752 &  0.654 &  0.690 &  0.682 &  0.741 &  0.737 \\
 all-MiniLM-L6-v2 + MCR100 & 0.778 &  0.685 &  0.742 &  0.721 &  0.780 &  0.777 \\
 all-MiniLM-L6-v2 + MCR200 & 0.810 &  0.705 &  0.773 &  0.751 &  0.813 &  0.792 \\
 all-MiniLM-L6-v2 + MCR300 & 0.813 &  0.710 &  0.780 &  0.759 &  0.826 &  0.800 \\
\textbf{all-MiniLM-L6-v2 (384)} & 0.824 &  0.711 &  0.790 &  0.772 &  0.838 &  0.812 \\
\bottomrule
\end{tabular}
\end{table*}

\begin{table*}[!h]
\scriptsize
    \caption{\small For Sentence Classification (SC) tasks backbone model achieves the best results (bolded) for accuracy on multiple benchmarks, although we observe comparable results of our method compared to the backbone and distilled model.}
    \label{tab:SC_results}
    \centering
\begin{tabular}{lrrrrrrr}
\toprule
                      model &  SST2 &  SST5 &    MR &    CR &  SUBJ &  MPQA &  TREC \\
\midrule
 all-mpnet-base-v2 + MCR50 & 75.45 & 36.43 & 69.67 & 63.76 & 79.16 & 68.84 &  51.6 \\
all-mpnet-base-v2 + MCR100 & 82.54 & 39.19 & 75.85 & 63.76 & 81.86 & 68.77 &  60.0 \\
all-mpnet-base-v2 + MCR200 & 86.55 & 42.76 & 80.62 & 72.77 & 88.28 & 82.27 &  71.0 \\
all-mpnet-base-v2 + MCR300 & 87.59 & 44.66 & 82.33 & 79.71 & 90.73 & 85.76 &  79.8 \\
         \textbf{all-mpnet-base-v2 (768)} & 88.74 & 49.00 & 85.05 & 86.84 & 93.97 & 89.32 &  94.0 \\
\midrule
  all-MiniLM-L6-v2 + MCR50 & 65.95 & 31.76 & 61.61 & 63.76 & 79.19 & 68.77 &  41.0 \\
 all-MiniLM-L6-v2 + MCR100 & 72.27 & 33.94 & 66.38 & 63.82 & 83.97 & 76.58 &  64.0 \\
 all-MiniLM-L6-v2 + MCR200 & 77.54 & 37.10 & 70.06 & 69.17 & 86.87 & 81.83 &  69.2 \\
 all-MiniLM-L6-v2 + MCR300 & 79.35 & 39.50 & 72.95 & 75.07 & 88.47 & 84.13 &  72.0 \\
          \textbf{all-MiniLM-L6-v2 (384)} & 81.44 & 42.99 & 75.98 & 80.56 & 91.80 & 87.38 &  90.0 \\
\bottomrule
\end{tabular}
\end{table*}

%% file: main.bbl
\begin{thebibliography}{10}

\bibitem{https://doi.org/10.48550/arxiv.1806.04713}
Hanan Aldarmaki and Mona Diab.
\newblock Evaluation of unsupervised compositional representations, 2018.

\bibitem{10.5555/944919.944966}
Yoshua Bengio, R\'{e}jean Ducharme, Pascal Vincent, and Christian Janvin.
\newblock A neural probabilistic language model.
\newblock {\em J. Mach. Learn. Res.}, 3(null):1137–1155, mar 2003.

\bibitem{bojanowski2016enriching}
Piotr Bojanowski, Edouard Grave, Armand Joulin, and Tomas Mikolov.
\newblock Enriching word vectors with subword information.
\newblock {\em arXiv preprint arXiv:1607.04606}, 2016.

\bibitem{bowman-etal-2015-large}
Samuel~R. Bowman, Gabor Angeli, Christopher Potts, and Christopher~D. Manning.
\newblock A large annotated corpus for learning natural language inference.
\newblock In {\em Proceedings of the 2015 Conference on Empirical Methods in
  Natural Language Processing}, pages 632--642, Lisbon, Portugal, September
  2015. Association for Computational Linguistics.

\bibitem{https://doi.org/10.48550/arxiv.1511.07289}
Djork-Arné Clevert, Thomas Unterthiner, and Sepp Hochreiter.
\newblock Fast and accurate deep network learning by exponential linear units
  (elus), 2015.

\bibitem{conneau-kiela-2018-senteval}
Alexis Conneau and Douwe Kiela.
\newblock {S}ent{E}val: An evaluation toolkit for universal sentence
  representations.
\newblock In {\em Proceedings of the Eleventh International Conference on
  Language Resources and Evaluation ({LREC} 2018)}, Miyazaki, Japan, May 2018.
  European Language Resources Association (ELRA).

\bibitem{conneau-etal-2017-supervised}
Alexis Conneau, Douwe Kiela, Holger Schwenk, Lo{\"\i}c Barrault, and Antoine
  Bordes.
\newblock Supervised learning of universal sentence representations from
  natural language inference data.
\newblock In {\em Proceedings of the 2017 Conference on Empirical Methods in
  Natural Language Processing}, pages 670--680, Copenhagen, Denmark, September
  2017. Association for Computational Linguistics.

\bibitem{10.5555/1146355}
Thomas~M. Cover and Joy~A. Thomas.
\newblock {\em Elements of Information Theory (Wiley Series in
  Telecommunications and Signal Processing)}.
\newblock Wiley-Interscience, USA, 2006.

\bibitem{devlin-etal-2019-bert}
Jacob Devlin, Ming-Wei Chang, Kenton Lee, and Kristina Toutanova.
\newblock {BERT}: Pre-training of deep bidirectional transformers for language
  understanding.
\newblock In {\em Proceedings of the 2019 Conference of the North {A}merican
  Chapter of the Association for Computational Linguistics: Human Language
  Technologies, Volume 1 (Long and Short Papers)}, pages 4171--4186,
  Minneapolis, Minnesota, June 2019. Association for Computational Linguistics.

\bibitem{gao-etal-2021-simcse}
Tianyu Gao, Xingcheng Yao, and Danqi Chen.
\newblock {S}im{CSE}: Simple contrastive learning of sentence embeddings.
\newblock In {\em Proceedings of the 2021 Conference on Empirical Methods in
  Natural Language Processing}, pages 6894--6910, Online and Punta Cana,
  Dominican Republic, November 2021. Association for Computational Linguistics.

\bibitem{hoogeveen2015}
Doris Hoogeveen, Karin~M. Verspoor, and Timothy Baldwin.
\newblock Cqadupstack: A benchmark data set for community question-answering
  research.
\newblock In {\em Proceedings of the 20th Australasian Document Computing
  Symposium (ADCS)}, ADCS '15, pages 3:1--3:8, New York, NY, USA, 2015. ACM.

\bibitem{Gumbel-Softmax}
Iris A.~M. Huijben, Wouter Kool, Max~B. Paulus, and Ruud J.~G. van Sloun.
\newblock A review of the gumbel-max trick and its extensions for discrete
  stochasticity in machine learning, 2021.

\bibitem{johnson2019billion}
Jeff Johnson, Matthijs Douze, and Herv{\'e} J{\'e}gou.
\newblock Billion-scale similarity search with {GPUs}.
\newblock {\em IEEE Transactions on Big Data}, 7(3):535--547, 2019.

\bibitem{10.5555/2969442.2969607}
Ryan Kiros, Yukun Zhu, Ruslan Salakhutdinov, Richard~S. Zemel, Antonio
  Torralba, Raquel Urtasun, and Sanja Fidler.
\newblock Skip-thought vectors.
\newblock In {\em Proceedings of the 28th International Conference on Neural
  Information Processing Systems - Volume 2}, NIPS'15, page 3294–3302,
  Cambridge, MA, USA, 2015. MIT Press.

\bibitem{li-etal-2020-sentence}
Bohan Li, Hao Zhou, Junxian He, Mingxuan Wang, Yiming Yang, and Lei Li.
\newblock On the sentence embeddings from pre-trained language models.
\newblock In {\em Proceedings of the 2020 Conference on Empirical Methods in
  Natural Language Processing (EMNLP)}, pages 9119--9130, Online, November
  2020. Association for Computational Linguistics.

\bibitem{li_neural_2022}
Zengyi Li, Yubei Chen, Yann LeCun, and Friedrich~T. Sommer.
\newblock Neural {Manifold} {Clustering} and {Embedding}, January 2022.
\newblock arXiv:2201.10000 [cs].

\bibitem{DBLP:journals/corr/abs-1803-02893}
Lajanugen Logeswaran and Honglak Lee.
\newblock An efficient framework for learning sentence representations.
\newblock {\em CoRR}, abs/1803.02893, 2018.

\bibitem{MCR2-volume-Proof}
Yi~Ma, Harm Derksen, Wei Hong, and John Wright.
\newblock Segmentation of multivariate mixed data via lossy data coding and
  compression.
\newblock {\em IEEE Transactions on Pattern Analysis and Machine Intelligence},
  29(9):1546--1562, 2007.

\bibitem{Mikolov13}
Tomas Mikolov, Ilya Sutskever, Kai Chen, Greg Corrado, and Jeffrey Dean.
\newblock Distributed representations of words and phrases and their
  compositionality, 2013.

\bibitem{Pagliardini_2018}
Matteo Pagliardini, Prakhar Gupta, and Martin Jaggi.
\newblock Unsupervised learning of sentence embeddings using compositional
  n-gram features.
\newblock In {\em Proceedings of the 2018 Conference of the North American
  Chapter of the Association for Computational Linguistics: Human Language
  Technologies, Volume 1 (Long Papers)}. Association for Computational
  Linguistics, 2018.

\bibitem{patel-bhattacharyya-2017-towards}
Kevin Patel and Pushpak Bhattacharyya.
\newblock Towards lower bounds on number of dimensions for word embeddings.
\newblock In {\em Proceedings of the Eighth International Joint Conference on
  Natural Language Processing (Volume 2: Short Papers)}, pages 31--36, Taipei,
  Taiwan, November 2017. Asian Federation of Natural Language Processing.

\bibitem{pennington-etal-2014-glove}
Jeffrey Pennington, Richard Socher, and Christopher Manning.
\newblock {G}lo{V}e: Global vectors for word representation.
\newblock In {\em Proceedings of the 2014 Conference on Empirical Methods in
  Natural Language Processing ({EMNLP})}, pages 1532--1543, Doha, Qatar,
  October 2014. Association for Computational Linguistics.

\bibitem{peters-etal-2018-deep}
Matthew~E. Peters, Mark Neumann, Mohit Iyyer, Matt Gardner, Christopher Clark,
  Kenton Lee, and Luke Zettlemoyer.
\newblock Deep contextualized word representations.
\newblock In {\em Proceedings of the 2018 Conference of the North {A}merican
  Chapter of the Association for Computational Linguistics: Human Language
  Technologies, Volume 1 (Long Papers)}, pages 2227--2237, New Orleans,
  Louisiana, June 2018. Association for Computational Linguistics.

\bibitem{reimers_sentence-bert_2019}
Nils Reimers and Iryna Gurevych.
\newblock Sentence-{BERT}: {Sentence} {Embeddings} using {Siamese}
  {BERT}-{Networks}, August 2019.
\newblock arXiv:1908.10084 [cs].

\bibitem{reimers-2020-multilingual-sentence-bert}
Nils Reimers and Iryna Gurevych.
\newblock Making monolingual sentence embeddings multilingual using knowledge
  distillation.
\newblock In {\em Proceedings of the 2020 Conference on Empirical Methods in
  Natural Language Processing}. Association for Computational Linguistics, 11
  2020.

\bibitem{williams-etal-2018-broad}
Adina Williams, Nikita Nangia, and Samuel Bowman.
\newblock A broad-coverage challenge corpus for sentence understanding through
  inference.
\newblock In {\em Proceedings of the 2018 Conference of the North {A}merican
  Chapter of the Association for Computational Linguistics: Human Language
  Technologies, Volume 1 (Long Papers)}, pages 1112--1122, New Orleans,
  Louisiana, June 2018. Association for Computational Linguistics.

\bibitem{yu_learning_2020}
Yaodong Yu, Kwan Ho~Ryan Chan, Chong You, Chaobing Song, and Yi~Ma.
\newblock Learning {Diverse} and {Discriminative} {Representations} via the
  {Principle} of {Maximal} {Coding} {Rate} {Reduction}, June 2020.
\newblock arXiv:2006.08558 [cs, math, stat].

\end{thebibliography}
